\documentclass[lettersize,journal]{IEEEtran}

\usepackage{array}
\usepackage[caption=false,font=normalsize,labelfont=sf,textfont=sf]{subfig}
\usepackage{textcomp}
\usepackage{stfloats}
\usepackage{url}
\usepackage{verbatim}
\usepackage{graphicx}
\usepackage{cite}
\def\BibTeX{{\rm B\kern-.05em{\sc i\kern-.025em b}\kern-.08em
        T\kern-.1667em\lower.7ex\hbox{E}\kern-.125emX}}
\usepackage{balance}

\usepackage{makecell}
\usepackage{booktabs}
\usepackage{threeparttable}
\usepackage{multirow}
\usepackage{tabularx}

\usepackage{amssymb}
\usepackage{bm}
\usepackage{mathtools}
\usepackage{nomencl}
\graphicspath{{fig/}, {author/}}

\usepackage{color, xcolor}
\usepackage{colortbl}
\usepackage{soul}
\soulregister{\cite}7       
\soulregister{\citep}7      
\soulregister{\citet}7      
\soulregister{\ref}7        
\soulregister{\pageref}7    
\sethlcolor{yellow}         

\usepackage{amsmath,amsfonts}
\usepackage{algorithmic}
\usepackage{algorithm}



\begin{document}

\title{Artificial Intelligence-Enabled Space Robot Operations: Technologies, Challenges and Prospects}

\author{Zeyuan Huang, Gang Chen, Zixuan Hao, Guoqin Tang, Junyi Zong, Guoyou Ban, Jiale Wang, Haoyang Lv, Chaoqian Ren, Sitong Liu

\thanks{
This work was supported by National Natural Science Foundation of China (62573064) and State Key Laboratory of Robotics and Systems (HIT) (SKLRS-2025-KF-07).
(Corresponding author: Gang Chen)

The authors are with the School of Intelligent Engineering and Automation, Beijing University of Posts and Telecommunications, Beijing, China (e-mail: huangzeyuan1994@163.com; buptcg@163.com; haozixuan2001@163.com; tgq@bupt.edu.cn; zongjunyi@bupt.edu.cn; 13953002921@163.com; jiale.wang@bupt.edu.cn; lvhy13834664292@bupt.edu.cn; rcq@bupt.edu.cn; sitongliu2026@163.com).
    
}}


\markboth{}{}


\maketitle


\begin{abstract}
Space robots are increasingly expected to perform long-duration, contact-rich, and multi-stage operations with limited human intervention. Recent advances in artificial intelligence (AI), robot learning, and embodied foundation models provide new opportunities to improve the autonomy and adaptability of such systems, but their transfer to space is constrained by scarce mission data, space-specific dynamics and sensing conditions, limited onboard resources, and stringent safety requirements. This article reviews artificial intelligence-enabled space robot operations (AI-SRO) from a capability-building perspective. We first summarize representative operational scenarios, autonomy trends, and space-specific constraints. We then establish a three-layer technical framework comprising capability foundations, capability formation, and capability deployment/evolution. Within this framework, we review simulation environments, datasets and benchmarks; task and environment understanding, state perception, decision-making and planning, and action execution; and onboard deployment, ground-to-space adaptation, continual learning, and capability transfer. Finally, we propose key research directions toward trustworthy simulation and data, open-world multimodal cognition, long-horizon safe decision-making, physically constrained policy learning, and space computing infrastructures.
\end{abstract}



\section{Introduction}
\label{sec_introduction}

\IEEEPARstart{W}{ith} the expansion of space-station construction and operation, on-orbit servicing, deep-space exploration, and extraterrestrial surface activities, space missions are becoming increasingly complex and long-duration. Space robots can assist or replace astronauts in hazardous environments and have become important systems for improving mission safety, efficiency, and sustainability \cite{Gao2017ReviewSpace_R001,FloresAbad2014ReviewSpace_R002}. Space robots refer to robotic systems deployed in orbital environments, inside or outside spacecraft, or on extraterrestrial surfaces, with capabilities for perception, motion control, and physical interaction. We use Space Robot Operations (SRO) to refer to mission-level robotic operations such as inspection, approach and capture, servicing and maintenance, assembly and construction, sampling, and equipment deployment. These operations are accomplished through a closed loop of task and environment understanding, state perception, decision-making and planning, and physical action execution.

SRO is moving beyond fixed platforms, standardized interfaces, and predefined procedures toward missions involving multiple targets, multiple stages, long operating periods, and cross-platform collaboration. However, current systems still rely heavily on accurate pre-mission models, ground-based planning, teleoperation or supervisory control, and task-specific programs \cite{Li2022SurveySpace_R003,Alizadeh2024ComprehensiveSurvey_R004}. Communication delay and interruption, unexpected target states, and changing mission procedures can therefore limit responsiveness and long-term autonomy \cite{Chen2026KeyTechnical_R005}.

Advances in imitation learning, reinforcement learning, and embodied foundation models such as Vision-Language-Action (VLA) models provide new means for robots to acquire perception, reasoning, planning, and manipulation capabilities from data and interaction \cite{ONeill2024OpenX_R006,Zitkovich2023RT2_R007,Kim2025OpenVLAOpen_R008,Chi2025DiffusionPolicy_R009}. In this review, we use the term Artificial Intelligence-enabled Space Robot Operations (AI-SRO) to describe SRO in which AI methods are integrated into the operational loop to enhance autonomy, adaptability, and generalization. Yet terrestrial AI methods cannot be transferred directly. Microgravity, reduced gravity, free-floating dynamics, and compliant contacts change robot behavior; extreme illumination, radiation, vacuum, and temperature variations challenge sensing, actuation, hardware reliability, and onboard computing; onboard computation, power, thermal dissipation, and communication are tightly constrained; and real space-operation data, especially anomalous and failure data, are scarce. Moreover, on-orbit trial and error is expensive and potentially hazardous. Therefore, AI-SRO must address learning, physical consistency, deployability, and safety as a coupled system problem.

Prior surveys have examined space robotics, on-orbit servicing and assembly, capture, manipulation, visual servoing, motion planning, and control \cite{Gao2017ReviewSpace_R001,FloresAbad2014ReviewSpace_R002,Li2022SurveySpace_R003,Alizadeh2024ComprehensiveSurvey_R004,Chen2026KeyTechnical_R005,Papadopoulos2021RoboticManipulation_R010,AmayaMejia2024VisualServoing_R011}. Most organize the literature around robot platforms, mission classes, or individual technical modules such as perception, planning, manipulation, and control. Comparatively less attention has been given to how AI capabilities are built, deployed, and evolved across the SRO lifecycle, from simulation and data foundations to capability formation and resource-constrained space deployment. Such a capability-oriented perspective is important because simulation fidelity and data coverage shape learned capability boundaries, while onboard resources and safety requirements constrain deployment and adaptation.

Accordingly, this article reviews AI-SRO through a three-layer framework of capability foundations, capability formation, and capability deployment and evolution. Its main contributions are:
\begin{itemize}
    \item We establish a capability-oriented three-layer framework for AI-SRO, linking capability foundations, capability formation, and capability deployment and evolution to organize fragmented research across the operational lifecycle.
    \item We systematically review key technologies across these three layers, covering simulation, data, and evaluation foundations; task and environment understanding, state perception, decision-making and planning, and action execution; and onboard deployment, ground-to-space adaptation, and capability evolution.
    \item We identify major limitations that hinder reliable autonomous space robot operations and discuss future directions in trustworthy simulation and data, multimodal cognition, safe long-horizon decision-making, physically constrained action generation, and space computing and continual capability evolution.
\end{itemize}

\begin{figure}[!t]
    \centering
    \includegraphics[width=\columnwidth]{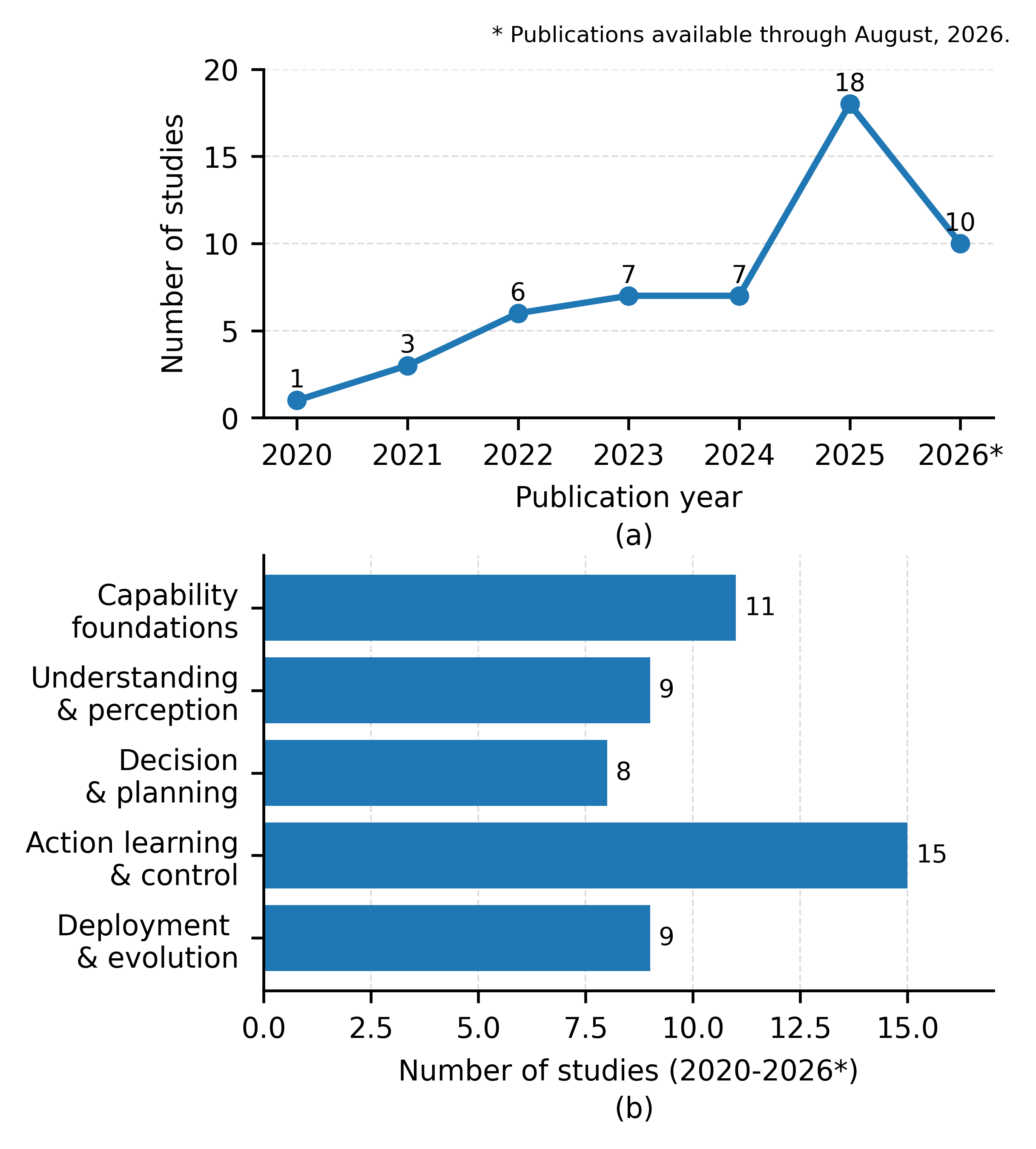}
    \caption{
        Distribution of recent AI-SRO studies included in this review: (a) annual publication trend from 2020 to 2026 and (b) distribution across major technical categories. The 2026 statistics include publications available through August, 2026.}
    \label{fig1}
\end{figure}

To clarify the survey scope, Fig.~\ref{fig1} summarizes recent AI-SRO studies published from 2020 to August 2026. Studies are counted as directly related to AI-SRO if they address space-robot operational tasks or dedicated datasets/benchmarks and contribute autonomous planning, learning-based, data-driven, or adaptive capabilities. Earlier foundational studies are discussed in the review but excluded from the trend statistics. Under these criteria, 52 studies are included. The number of AI-SRO studies increases markedly after 2022, with research concentrated mainly on action learning and control, while other areas remain comparatively underrepresented.

\section{Overview of Current Space Robot Operations}
\label{sec_overview}

Space robot operations vary substantially with the supporting platform, environment, and task object. Existing systems can be broadly grouped into space-station extravehicular operations, space-station intravehicular operations, on-orbit operations from satellite platforms, and extraterrestrial-surface operations. Table \ref{table_1} summarizes the four major SRO scenarios, together with representative systems, typical tasks, and their current operational characteristics.

\begin{table*}[t]
    \centering
    \caption{Representative SRO Scenarios and Current Operational Characteristics}
    \label{table_1}
    \footnotesize
    \renewcommand{\arraystretch}{1.08}
    \renewcommand{\tabularxcolumn}[1]{m{#1}}
    \begin{tabularx}{\textwidth}{|>{\raggedright\arraybackslash}m{2.65cm}|>{\raggedright\arraybackslash}X|>{\raggedright\arraybackslash}X|>{\raggedright\arraybackslash}X|}
    \hline
        \multicolumn{1}{|c|}{\textbf{Scenario}} &
        \multicolumn{1}{c|}{\textbf{Representative Systems}} &
        \multicolumn{1}{c|}{\textbf{Typical Tasks}} &
        \multicolumn{1}{c|}{\textbf{Current Operational Characteristics}} \\ \hline

        Space-station extravehicular
        & Canadarm2/Dextre \cite{Coleshill2009DextreImproving_R012,Aziz2013DevelopmentVerification_R013}; China Space Station manipulators \cite{Li2015OverviewChinese_R014}; GITAI S2 \cite{ISS2024AutonomousRobotic_R015}
        & Payload handling, module relocation, ORU servicing, inspection, and maintenance
        & Ground/crew-supervised and procedure-driven; mature for structured tasks, but unexpected states generally require human intervention \\ \hline

        Space-station intravehicular
        & Tiangong-2 manipulator \cite{LIU2018KeyTechnologies_R016}; GITAI S1 \cite{GITAI2021GITAIAutonomous_R018}
        & Object capture, tool use, switch/cable operations, assembly, and maintenance assistance
        & Teleoperation or human--robot collaboration with local autonomy; mainly supports predefined objects, interfaces, and procedures \\ \hline

        On-orbit servicing
        & ETS-VII \cite{Oda1994OrbitExperiment_R019}; Orbital Express \cite{Ma2023AdvancesSpace_R021}
        & Rendezvous and proximity operations, inspection, capture, servicing, and module replacement
        & Ground-planned with supervised/local autonomous execution; adaptation remains limited for noncooperative targets, uncertain dynamics, and unexpected contact \\ \hline

        Extraterrestrial surface
        & Phoenix \cite{Arvidson2009ResultsFrom_R022}, Curiosity \cite{Jandura2010OverviewMars_R023}, and Perseverance \cite{Townsend2022RoboticsVerification_R024}; Chang'e-5/6 \cite{Zhou2022ScientificObjectives_R025,HU2025OverallDesign_R026}
        & Excavation, drilling, sampling, sample handling, and instrument deployment
        & Ground-generated mission sequences with onboard local autonomy; communication delay favors local execution, but terrain and material uncertainty constrain adaptation \\ \hline
    \end{tabularx}
\end{table*}

\subsection{Typical SRO Scenarios and Tasks}

\textbf{(1) Extravehicular operations on space stations.}
Representative systems include Canadarm2 and Dextre for station assembly, payload handling, and orbital replacement unit maintenance \cite{Coleshill2009DextreImproving_R012,Aziz2013DevelopmentVerification_R013}; the China Space Station robotic manipulators for payload transfer and dexterous operations \cite{Li2015OverviewChinese_R014}; and GITAI S2 for servicing, assembly, and manufacturing demonstrations \cite{ISS2024AutonomousRobotic_R015}. These systems provide mature mobility and manipulation for structured equipment, but still depend strongly on standardized interfaces, specialized tools, pre-mission planning, and ground supervision.

\textbf{(2) Intravehicular operations on space stations.}
Intravehicular robots operate in confined, equipment-dense environments and must safely perform object capture, tool use, panel or switch manipulation, cable operations, assembly, and maintenance assistance around astronauts. Tiangong-2 demonstrated free-floating object capture, tool operation, teleoperation, and human--robot collaborative maintenance \cite{LIU2018KeyTechnologies_R016,Liu2019RoboticHand_R017}, while GITAI S1 demonstrated assembly, switch manipulation, and cable operations on the ISS \cite{GITAI2021GITAIAutonomous_R018}. Current autonomy is mainly local and procedure-driven, with limited support for open-ended instructions and unexpected situations.

\textbf{(3) On-orbit operations from satellite platforms.}
Servicing spacecraft equipped with manipulators can perform rendezvous, inspection, capture, maintenance, resupply, module replacement, and debris disposal. ETS-VII demonstrated autonomous rendezvous, docking, manipulation, and servicing \cite{Oda1994OrbitExperiment_R019}, while subsequent missions have explored robotic deployment, proximity servicing, and noncooperative-target capture \cite{Li2019OrbitService_R020,Ma2023AdvancesSpace_R021}. Because target states and interfaces may be unknown and contact strongly couples the servicer, manipulator, and target, these missions require continuous perception, motion estimation, safe approach, compliant capture, and post-capture stabilization.

\textbf{(4) Operations on extraterrestrial surfaces.}
Surface operations include drilling, sampling, scientific-instrument deployment, resource utilization, and construction. Mars systems such as Phoenix, Curiosity, and Perseverance have used manipulators for excavation, drilling, close-range science, and sample handling \cite{Arvidson2009ResultsFrom_R022,Jandura2010OverviewMars_R023,Townsend2022RoboticsVerification_R024}; Chang'e-5 and Chang'e-6 completed lunar sampling, transfer, and encapsulation \cite{Zhou2022ScientificObjectives_R025,HU2025OverallDesign_R026}. Ground planning remains important, while complex terrain, granular media, reduced gravity, long communication delay, and uncertain material properties increase demands on local adaptation and contact autonomy.

\subsection{Evolution of SRO Operational Paradigms}

Current SRO generally combines teleoperation, predefined autonomous procedures, and task-specific onboard autonomy. Near-Earth systems retain humans for high-level decisions and critical state confirmation, whereas planetary and deep-space robots execute short-timescale motion and manipulation locally from ground-generated mission sequences. This architecture provides controllability but leaves an autonomy gap: objects and interfaces are usually modeled in advance; perception, planning, and control are customized for individual tasks; deviations in target pose, dynamics, or contact conditions often require ground reassessment; and autonomous behaviors remain local and short-horizon. Future missions instead require closed-loop autonomy spanning task interpretation, perception, decision-making, physical execution, and recovery, motivating the introduction of AI and embodied intelligence.

\subsection{Space-Specific Constraints on AI-SRO}

AI-SRO is subject to four coupled constraints. Mission constraints arise from long-horizon, multistage, and high-risk procedures, requiring task decomposition, transition management, and replanning rather than isolated learned skills. Environmental and physical constraints include microgravity or reduced gravity, free-floating-base coupling, flexible structures, extreme illumination, radiation, vacuum, and thermal variation, which require physically consistent learning and uncertainty-aware perception. System-resource constraints include delayed or intermittent communication and strict limits on onboard computation, storage, power, thermal dissipation, and radiation tolerance, making local inference and predictable real-time performance essential. Embodiment and data constraints arise from the diversity of manipulators, free-flyers, and surface robots together with the scarcity of real mission and failure data, motivating high-fidelity simulation, analogue experiments, transfer learning, and physical priors.

These constraints distinguish AI-SRO from terrestrial learning settings based on abundant data, computing, and repeated trial-and-error interaction. They also motivate the capability-oriented framework introduced next.

\section{Overall Technical Framework for AI-Enabled Space Robot Operations}
\label{sec_framework}

\subsection{From Terrestrial Robot Learning to AI-SRO}

Modern robot learning is dominated by complementary paradigms. Imitation learning maps observations and states to actions from expert demonstrations and is effective when interactive data collection is expensive \cite{Argall2009SurveyRobot_R027}; reinforcement learning optimizes policies through environmental interaction and reward feedback, providing adaptability for complex sequential decisions \cite{Kober2013ReinforcementLearning_R028}. More recently, embodied foundation models including Vision-Language Models (VLMs), Vision-Language-Action models (VLAs), and World Action Models (WAMs), have used large-scale multimodal pretraining to support open-vocabulary perception, task reasoning, and cross-task policy generalization \cite{Zhang2024VisionLanguage_R029,Ma2026SurveyVision_R030,Wang2026WorldAction_R031}. Despite different learning mechanisms, these paradigms share a common process: resources and evaluation environments are constructed, models learn task-to-action mappings, and trained capabilities are adapted and deployed on physical robots. This common capability-building process motivates the lifecycle-oriented framework adopted for AI-SRO in this review.

\subsection{Three-Layer Technical Framework for AI-SRO}

\begin{figure*}[t]
    \centering
    \includegraphics[width=\textwidth]{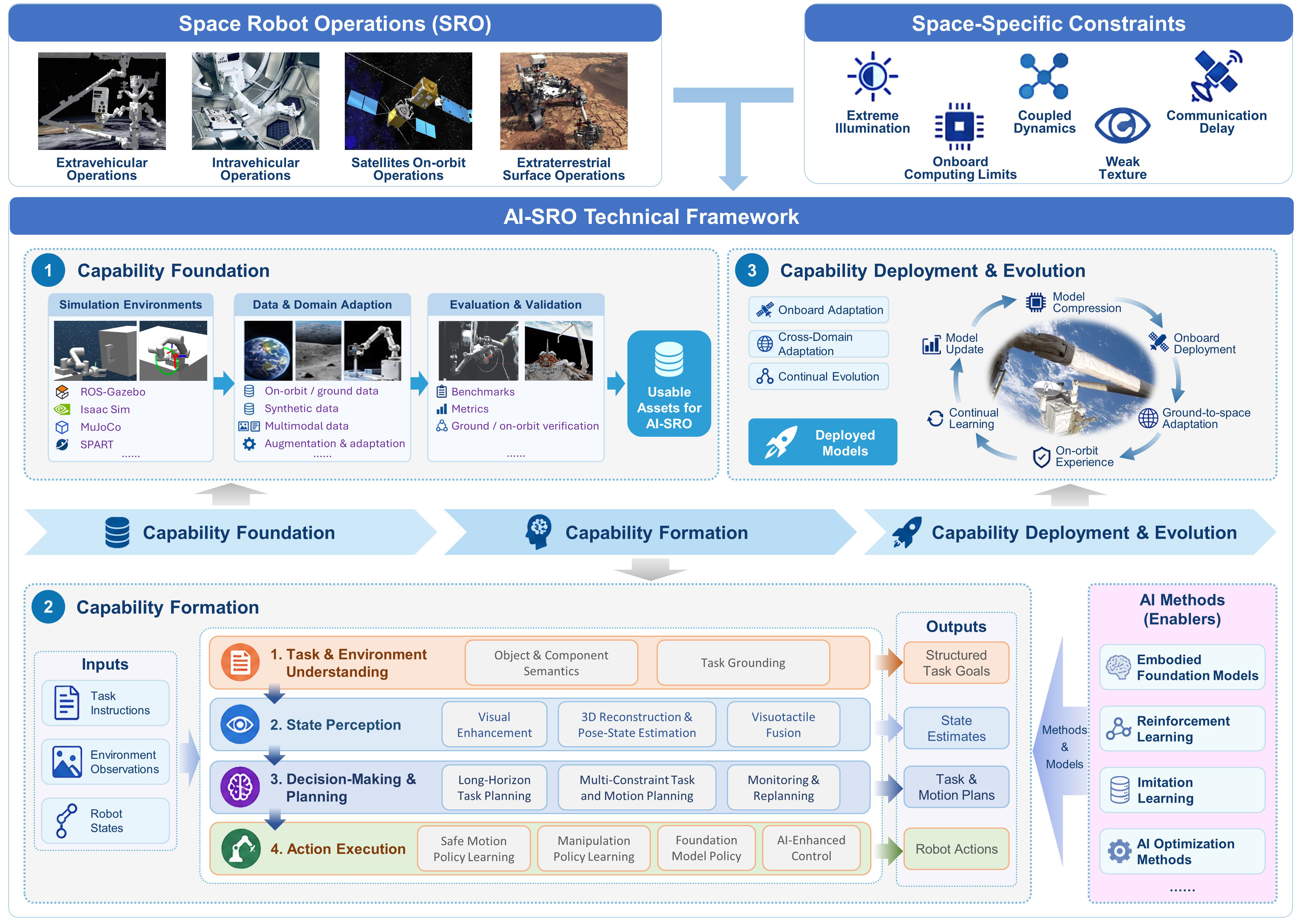}
    \caption{Technical framework of AI-enabled space robot operations (AI-SRO). Under representative SRO scenarios and space-specific constraints, the framework comprises three interconnected layers: capability foundation, capability formation, and capability deployment and evolution. AI methods provide cross-cutting support for transforming task, environment, and robot-state information into executable capabilities and for their subsequent onboard deployment, adaptation, and continual evolution.}
    \label{fig2}
\end{figure*}

Because space-specific mission, physical, data, and resource constraints affect every stage of this process, this article organizes AI-SRO using a three-layer framework comprising capability foundations, capability formation, and capability deployment and evolution, as illustrated in Fig.~\ref{fig2}.

\textbf{(1) Capability foundations} comprise simulation environments, multi-source datasets and data generation, cross-domain augmentation, and evaluation benchmarks. They reproduce space scenes, robot dynamics, contacts, anomalous states, and task distributions, thereby defining the data and validation envelope within which intelligent capabilities are learned.

\textbf{(2) Capability formation} establishes the closed loop from mission intent to physical interaction. Task and environment understanding grounds instructions and object semantics; state perception estimates geometric, motion, and contact states; decision-making and planning generate task sequences and feasible motions; and action execution converts these results into safe behavior while execution feedback updates subsequent perception and planning.

\textbf{(3) Capability deployment and evolution} addresses the transition from ground training to long-duration space operation through resource-constrained onboard inference, ground-to-space adaptation and calibration, continual learning, and transfer across tasks and embodiments. Progressive safety validation connects this layer with the evaluation infrastructure established in the capability foundations.

The three layers are bidirectionally coupled. Simulation fidelity and data coverage bound what models can learn, whereas onboard compute, control-cycle, and safety constraints feed back into model architecture and training choices. Conversely, on-orbit observations, anomalies, and execution outcomes can be returned to datasets, simulators, and benchmarks for calibration and capability expansion. Physical, data, resource, and safety constraints therefore span the complete AI-SRO lifecycle.

\section{AI-SRO Capability Foundation}
\label{sec_foundation}

AI-SRO capability formation depends on simulation environments that reproduce space-mission characteristics, data resources covering diverse operational states, and progressive evaluation and validation. Because real on-orbit and surface experiments are limited and anomalous samples are scarce, these resources provide the controllable and repeatable foundation required for training and validating task understanding, perception, planning, and action execution.

\begin{table*}[t]
    \centering
    \caption{Comparison of Simulation Platforms for Space Robot Operations}
    \label{table_2}
    \footnotesize
    \renewcommand{\arraystretch}{1.05}
    \renewcommand{\tabularxcolumn}[1]{m{#1}}
    
    \begin{tabularx}{\textwidth}{|m{2.2cm}|X|X|X|X|}
    \hline
        \multicolumn{1}{|c|}{\textbf{Category}} &
        \multicolumn{1}{c|}{\textbf{Project}} &
        \multicolumn{1}{c|}{\textbf{Main Advantages}} &
        \multicolumn{1}{c|}{\textbf{Main Limitations}} &
        \multicolumn{1}{c|}{\textbf{Functions}} \\ \hline

        ROS-Gazebo Integrated Envs
        & ROS-Gazebo \cite{quigley2009ros_R031p, koenig2004design_R031p2}, Astrobee \cite{NASAndAstrobeeRobot_R032}, Int-Ball2 \cite{Japan2023IntBall2_R033}, Space ROS \cite{OpenndSpaceROS_R034}
        & Mature interfaces and convenient integration of perception, planning, and control modules
        & Limited fidelity in visual simulation and space dynamics
        & Task scripting, software-interface testing, closed-loop functional validation \\ \hline

        Visual Simulation
        & Isaac Sim \cite{NVIDIAndNVIDIAIsaac_R036}, Space Robotics Bench \cite{Orsula2025SpaceRobotics_R037},  GRADE \cite{Bonetto2026GRADEGenerating_R038}
        & Physically based rendering, procedural generation, domain randomization, and parallel training
        & Limited accuracy in spacecraft dynamics simulation
        & Perception training, synthetic data generation, policy learning \\ \hline

        Contact-Interaction Simulation
        & MuJoCo envs \cite{Todorov2012MuJoCoPhysics_R039, Cao2023ReinforcementLearning_R040,Peng2024ReinforcementLearning_R041,Zhao2024SpaceoctopusOctopus_R042}
        & Efficient contact solving and suitability for high-frequency policy interaction
        & Limited modeling capability for flexible bodies and complex impacts
        & Grasping, assembly, capture, reinforcement learning \\ \hline

        Spacecraft Dynamics Simulation
        & SPART \cite{VirgiliLlop2016SpacecraftRobotics_R043}, 42 \cite{NASAnd42General_R044}, Basilisk \cite{Kenneally2020BasiliskFlexible_R045}, Trick \cite{Penn2016TrickSimulation_R046}
        & High-fidelity modeling of orbital and attitude dynamics, multibody coupling, and GNC systems
        & Relatively limited support for visual simulation, contact interaction, and learning interfaces
        & Dynamics analysis, control validation, system-level simulation \\ \hline

        Space-Imaging Simulation
        & PANGU \cite{Martin2019PlanetarySurface_R047}, SurRender \cite{Brochard2018ScientificImage_R048}
        & Strong capabilities in ray tracing and spaceborne sensor modeling
        & Does not directly support space robotic manipulation simulation
        & Navigation, target recognition, imaging-pipeline validation \\ \hline
    \end{tabularx}
\end{table*}

\subsection{Simulation Environments for Space Robot Operations}

Space robot simulation involves more than rigid-body motion or visual-scene simulation. Depending on the stage of algorithm development, it must support different requirements, including closed-loop task integration, visual data generation, contact-interaction training, spacecraft dynamics modeling, and validation of specialized imaging pipelines. Different platforms offer distinct advantages in software interfaces, image fidelity, dynamics accuracy, and computational efficiency. Consequently, co-simulation across multiple platforms is often required to jointly support AI-SRO capability formation.

\textbf{(1) ROS-Gazebo integrated environments.}
Space robot operations involve multiple modules, including perception, planning, control, and task scheduling. Simulation platforms must therefore support the integrated operation of task scripts, robotic middleware, sensor interfaces, and controllers. Integrated environments based on ROS and Gazebo provide open interfaces, convenient module reuse, and strong closed-loop integration capabilities\cite{quigley2009ros_R031p, koenig2004design_R031p2}. Astrobee \cite{NASAndAstrobeeRobot_R032} provides open-source software and a simulation environment for free-flying robots operating inside the International Space Station, supporting navigation, task execution, and closed-loop perception-control validation. Int-Ball2 \cite{Japan2023IntBall2_R033} provides a simulation environment corresponding to its flight software. Space ROS \cite{OpenndSpaceROS_R034} further adapts the ROS 2 ecosystem to the software quality, safety, and reusability requirements of space missions.

\textbf{(2) High-fidelity visual simulation.}
Extreme illumination variations, highly reflective surfaces, weak textures, and complex backgrounds can substantially reduce the reliability of space-vision models. Large-scale training samples therefore need to be generated using physically based rendering, procedural scene generation, and domain randomization. Isaac Sim \cite{NVIDIAndNVIDIAIsaac_R036} can integrate models of robots, cameras, materials, and physical environments, while supporting controllable synthetic-data generation as well as software-in-the-loop and hardware-in-the-loop testing. Building on this platform, Space Robotics Bench \cite{Orsula2025SpaceRobotics_R037} organizes space-robot models, procedurally generated scenes, parallel training, and benchmark tasks within a unified framework. GRADE \cite{Bonetto2026GRADEGenerating_R038} likewise uses Isaac Sim to construct dynamic, randomized, and highly realistic environments for large-scale data generation in visual perception and robot learning. Such platforms are well suited to offline and online training of visual models, although dedicated models are still required to capture space-specific optical effects and spacecraft dynamics.

\textbf{(3) Contact-interaction training simulation.}
Grasping, insertion, docking, assembly, and noncooperative-target capture require high-frequency computation of contact, friction, collision, and joint dynamics. MuJoCo \cite{Todorov2012MuJoCoPhysics_R039} employs efficient multibody dynamics and contact-solving methods, making it suitable for large-scale policy interaction and rapid iterative training. Dedicated simulation environments have been developed on this platform for motion planning of dual-arm free-floating robots, coordinated pose-based capture of noncooperative targets, and distributed policy learning for multi-arm robotic systems \cite{Cao2023ReinforcementLearning_R040, Peng2024ReinforcementLearning_R041, Zhao2024SpaceoctopusOctopus_R042}. Such platforms can improve the sample-generation efficiency of algorithms such as reinforcement learning. However, accurate simulation of flexible structures, complex materials, and high-speed impacts still depends on careful parameter identification and validation against higher-fidelity models.

\textbf{(4) Spacecraft dynamics simulation.}
The motion of space robots is affected by manipulator-base coupling, orbital and attitude dynamics, momentum exchange, and vibrations of flexible structures. SPART \cite{VirgiliLlop2016SpacecraftRobotics_R043} supports multibody dynamics and control modeling for space robots and enables algorithm validation in MATLAB/Simulink. The open-source spacecraft dynamics simulator 42, developed by NASA \cite{NASAnd42General_R044}, supports simulations of spacecraft attitude, orbital motion, environmental effects, and multi-spacecraft missions. Basilisk \cite{Kenneally2020BasiliskFlexible_R045} integrates dynamics models with guidance, navigation, and control modules through a modular software architecture. Trick \cite{Penn2016TrickSimulation_R046} supports time integration, data logging, fault injection, and distributed simulation for complex physical systems. These platforms are well suited to establishing physics-based reference models and verifying system stability, although their visual rendering and contact-learning capabilities are generally limited.

\textbf{(5) Specialized space-imaging simulation.}
General robot simulators struggle to fully reproduce, at the level of physical causation, the incidence of sunlight, surface reflectance, deep-space background, camera distortion, and sensor noise. PANGU \cite{Martin2019PlanetarySurface_R047} is designed for planetary surfaces such as the Moon, Mars, and asteroids, and supports multiresolution terrain modeling and planetary-surface image generation. SurRender \cite{Brochard2018ScientificImage_R048} employs physically based ray tracing and sensor modeling to generate high-fidelity images and depth data for spacecraft, planetary, and small-body scenarios. These tools primarily support visual navigation, target recognition, and imaging-pipeline validation, and generally need to be integrated with spacecraft dynamics models and space-robot controllers.

Overall, existing tools can separately support mission software, visual simulation, contact interaction, and spacecraft dynamics. However, a unified platform that simultaneously provides physical consistency, high-fidelity imaging, parallel learning, and closed-loop task execution remains lacking. Inconsistencies in model formats, time synchronization, and data interfaces among different simulators further increase the difficulty of multi-platform joint validation and scenario reuse.

\subsection{Dataset Construction and Cross-Domain Generation}

\textbf{(1) Data Sources and Dataset Construction}

Data for space robots can be obtained from real on-orbit or surface missions, ground-based analogue experiments, synthetic simulation, and Internet-scale or general-purpose robotic datasets. Data collected from real space missions provide the highest environmental fidelity, but are limited in quantity, concentrated in a small number of scenarios, and difficult to annotate. For example, the public Astrobee dataset contains images, inertial measurements, and pose information collected inside the International Space Station and can be used for research on free-flying robot navigation \cite{Kang2024AstrobeeISS_R049}. Ground-based analogue and hardware-in-the-loop data can reproduce selected space conditions through controlled illumination, target mock-ups, and motion platforms. SPEED+ provides both synthetic-domain and hardware-in-the-loop-domain data for evaluating cross-domain performance in spacecraft pose estimation \cite{Park2022SPEEDNext_R050}. Internet-scale and general-purpose robotic datasets can additionally provide object semantics, task knowledge, and basic manipulation priors. However, their environmental, embodiment, and action distributions differ substantially from those of SRO, making them difficult to use directly for training low-level action policies.

Synthetic simulation data offer the advantages of large scale, relatively low generation cost, and complete ground-truth annotations, and have therefore become a major source of training data for space perception models. ALLO establishes a high-fidelity image and data-generation pipeline for anomaly detection in close-proximity robotic operations around the Moon \cite{Leveugle2026PhotorealisticDataset_R051}. SPARK provides RGB and depth data under different orbital conditions, viewpoints, and illumination settings for spacecraft and space-debris recognition \cite{Musallam2021SpacecraftRecognition_R052}. Hoang et al. developed a spacecraft dataset containing bounding boxes, instance-segmentation masks, and component-level labels \cite{Hoang2021SpacecraftDataset_R053}. NCSTP further provides target-category, key-component, and multitask perception annotations for noncooperative space targets \cite{Liu2025LargeScale_R054}. SpaceDet incorporates orbital motion and camera-imaging characteristics into its data-generation process to support space-object detection research \cite{Xiao2025SpaceDetLarge_R055}. SPEED and SPEED+, meanwhile, have become widely used training and cross-domain evaluation resources for six-degree-of-freedom spacecraft pose estimation \cite{Kisantal2020SatellitePose_R056, Park2022SPEEDNext_R050}.

\textbf{(2) Multimodal and Multi-Granularity Annotation}

Complex SRO tasks often require simultaneous acquisition of RGB images, depth maps, point clouds, robot joint states, target poses, force measurements, and tactile information. Datasets should therefore not only ensure temporal synchronization and coordinate consistency across multiple sensing modalities, but also provide annotations at multiple levels of granularity, ranging from object-level and component-level labels to pixel-level and physical-state annotations. SpaceSense-Bench generates synchronized RGB images, depth maps, LiDAR point clouds, semantic labels, and six-degree-of-freedom pose ground truth in simulation, providing a unified data foundation for object detection, segmentation, and pose estimation \cite{Wu2026SpaceSenseBench_R057}. The datasets developed by Hoang et al. and NCSTP further emphasize spacecraft-instance and component-level semantic annotations, respectively \cite{Hoang2021SpacecraftDataset_R053, Liu2025LargeScale_R054}. Such data can support a continuous perception pipeline from object recognition and component parsing to manipulation-region localization. However, existing public resources remain predominantly vision-oriented and generally lack joint-state, force, tactile, contact-state, and failure-recovery data covering complete manipulation processes.

\textbf{(3) Data Augmentation and Cross-Domain Fusion}

Synthetic data and real space observations generally differ in material appearance, illumination, noise characteristics, and dynamic distributions. To reduce this domain gap, the data-generation process can randomize target dimensions and poses, backgrounds, solar incidence directions, surface materials, and camera parameters, while injecting degradations such as overexposure, shadows, noise, distortion, and radiation-induced interference. NCSTP expands sample diversity through randomization of targets, backgrounds, and illumination, whereas SpaceDet incorporates orbital dynamics and camera-noise models into its generation pipeline \cite{Liu2025LargeScale_R054, Xiao2025SpaceDetLarge_R055}. Building on these approaches, domain adaptation, feature-distribution alignment, image-style transfer, and fine-tuning with a small amount of real data can be used to integrate data from simulation, ground-test, and on-orbit domains. SPEED+ includes synthetic, optical laboratory, and hardware-in-the-loop domains, enabling quantitative evaluation of cross-domain model performance \cite{Park2022SPEEDNext_R050}. Compared with simply pursuing visual similarity, data generation for AI-SRO should additionally preserve the physical consistency among target motion, contact states, and sensor degradation, thereby avoiding augmented samples that violate actual space-environment dynamics or sensing characteristics.

\subsection{Validation and Evaluation Benchmarks}

Evaluation of AI-SRO should cover model-level performance, task-level operational capability, and cross-domain generalization. At the module level, evaluation metrics may include object-detection accuracy, intersection over union for segmentation, pose-estimation error, state-tracking error, planning success rate and computation time, trajectory-tracking error, and contact-force error. At the task level, evaluation should focus on task success rate, completion time, numbers of collisions and anomalies, energy and computational resource consumption, and generalization to unseen objects, environments, and tasks. For high-risk missions, worst-case performance, uncertainty calibration, and stability over repeated executions should also be reported, rather than relying solely on average success rates.

To reduce the risks associated with directly introducing AI models into space missions, AI-SRO generally adopts a progressive validation chain consisting of digital simulation, ground-based robotic experiments, space-analogue environment tests, software-in-the-loop or hardware-in-the-loop testing, and finally on-orbit or surface validation. Algorithms are first verified in controlled environments, after which task diversity and anomaly coverage are gradually expanded, while the fidelity of robot hardware, sensors, dynamics, communication conditions, and computing resources is progressively increased. Chapin et al. \cite{Chapin2025AutonomousPlanning_R058} designed a series of autonomous planning and assembly tests for the Astrobee robot, illustrating a progressive pathway from algorithm development and ground validation to deployment on an actual space platform. Throughout this process, each stage should employ traceable scenarios, initial conditions, and evaluation metrics so that changes in ground-test performance can be mapped to on-orbit risk, while failure cases identified during validation are fed back into simulation environments, datasets, and training pipelines.

Existing benchmarks are mostly designed for specific technical components and are commonly evaluated through simulation or ground experiments. SPEED and SPEED+ primarily evaluate spacecraft pose estimation and cross-domain generalization \cite{Kisantal2020SatellitePose_R056, Park2022SPEEDNext_R050}. Based on the SpaceBot Cup, Insaurralde et al. \cite{Insaurralde2016BenchmarkingAssessment_R059} proposed a technology-readiness assessment method for space robots that evaluates robotic systems in terms of mobility, manipulation, and autonomy. Orsula et al. \cite{Orsula2024TowardsBenchmarking_R060} developed a benchmarking methodology for robotic manipulation capabilities involving capture, sampling, insertion, and assembly tasks. Space Robotics Bench provides multiple learning tasks and evaluation environments for space robots \cite{Orsula2025SpaceRobotics_R037}. NASA Space Robotics Challenge 2 evaluates autonomous navigation, task planning, and manipulation capabilities of multi-robot systems in scenarios involving lunar resource prospecting, transportation, and collaborative operations \cite{Kilic2021NASASpace_R061}. These efforts have promoted a gradual transition from evaluating individual algorithmic metrics toward assessing task-level autonomous operational capabilities. Nevertheless, perception, planning, control, and failure recovery are still typically evaluated separately, and a unified benchmark covering the complete closed-loop SRO process has yet to be established.

Overall, current resources remain fragmented, with insufficient real and multimodal manipulation data, significant simulation-to-reality gaps, and inconsistent task-level evaluation. Reusable digital scenarios, standardized multimodal formats, and benchmarks spanning nominal, anomalous, and failure states are therefore needed to support reliable AI-SRO training and deployment.


\section{AI-SRO Capability Formation}
\label{sec_formation}

Supported by the simulation environments, data resources, and evaluation infrastructures described in Section \ref{sec_foundation}, the core objective of AI-SRO capability formation is to establish mappings among task instructions, environmental observations, robot states, and physical actions, enabling space robots to autonomously understand tasks, perceive states, plan operational processes, and generate executable actions. According to the basic information flow involved in complex space robot operations, this process can be divided into four stages: task and environment understanding, state perception, decision-making and planning, and action execution. Although these stages have distinct functional roles, they form a tightly coupled closed loop through state information, actions, and execution feedback. Even as large-model approaches such as VLAs increasingly blur the boundaries among conventional modules, these four stages remain fundamental dimensions for analyzing model functionality, performance bottlenecks, and safety risks.

\subsection{Task and Environment Understanding}

Task and environment understanding refers to the process of transforming task instructions and environmental information into logical representations and executable objectives for space robots. It directly determines whether a robot can correctly interpret its operating environment and understand mission intent. This process mainly consists of two interconnected stages. First, object and component semantic understanding is used to extract the categories, functions, and manipulation-related attributes of environmental entities and their components, thereby establishing a semantic representation grounded in the physical world. Subsequently, instruction understanding and task grounding map abstract task intentions to specific operational objects, manipulation regions, and action requirements, providing structured inputs for state perception, decision-making and planning, and action execution.

\textbf{(1) Object and Component Semantic Understanding}

Object and component semantic understanding is primarily responsible for extracting task-relevant semantic attributes of environmental entities, enabling space robots to identify operational objects, understand component functions, and determine their manipulability. Traditional space vision systems are generally designed for predefined spacecraft, tools, and interfaces. However, unstructured and open-set targets, such as space debris, satellite components, and natural objects on extraterrestrial surfaces, exhibit substantial variations in shape, size, and material properties, making accurate and comprehensive prior models difficult to obtain. Moreover, different space robot embodiments and end effectors possess different manipulation capabilities. Consequently, the semantic properties of a target are strongly coupled with task requirements, environmental states, and the morphology and capabilities of the robot itself.

On this background, existing research has increasingly progressed along the direction of open-set object retrieval, local component parsing, and manipulation-attribute understanding. Wang et al. proposed LE-Object \cite{Wang2025LEObject_R062}, which embeds language features into object-level neural radiance fields, enabling robots to retrieve previously unknown objects in three-dimensional scenes based on textual descriptions and providing a new technical route for open-vocabulary recognition of space targets. Zhu et al. \cite{Zhu2025ReviewAdvances_R063} noted that combining vision-language models with explicit or implicit 3D representations can facilitate the construction of spatial environment models containing both geometric and semantic information. Barad et al. \cite{Barad2024ObjectCentric_R064} employed 3D Gaussian splatting to jointly represent the geometry, appearance, and motion of dynamically unknown targets, providing a basis for component parsing and risk-region annotation of noncooperative spacecraft.

For complex SRO, merely identifying the overall category of a spacecraft or space facility is insufficient to support subsequent manipulation. Robots must further recognize components such as solar arrays, antennas, panels, connectors, and orbital replacement units, while distinguishing grasping interfaces, fastening locations, permissible contact regions, fragile components, and hazardous areas. Space-target understanding must therefore progress from object-level classification toward component-level semantic parsing and manipulation-attribute annotation, establishing relationships among objects, components, functions, and manipulation modes. Although existing methods have demonstrated preliminary capabilities for open-set retrieval and 3D semantic modeling of unknown objects, maintaining semantic consistency under severe illumination variations, highly dynamic motion, and substantial occlusion remains challenging. Furthermore, integrated representations of component functionality, physical properties, and manipulation risks are still insufficient.

\textbf{(2) Task Grounding}

Task grounding is responsible for transforming task intentions expressed in natural-language commands, mission scripts, operating procedures, and visual observations into operational objectives and action requirements corresponding to the current physical environment. This process must determine not only what task should be completed, but also which object and which part should be acted upon, what operation should be performed, and what preconditions, constraints, and completion criteria apply. Task grounding for AI-SRO faces three major challenges. First, commands transmitted from the ground are typically specified at the mission level, whereas space robots must combine these high-level instructions with local environmental observations and autonomously refine them into executable subtasks and action sequences. Second, the pose, visible regions, and manipulation windows of noncooperative targets may change rapidly, causing the initially established correspondence between tasks and objects to become invalid during execution. Third, delays and limited communication windows in space-to-ground links make frequent instruction confirmation and correction difficult, requiring space robots to possess a degree of autonomous semantic interpretation and state-updating capability.

To address the transformation from high-level missions to concrete actions, Jin et al. \cite{Jin2025LLMBased_R065} proposed a large language model-based approach for proximity-operation planning around noncooperative spacecraft. The method exploits spacecraft structural knowledge encoded in the model to supplement missing environmental priors under partial observations and transforms high-level operational objectives into sequences of active observation and close-proximity operations. Such studies indicate that large language models can exploit linguistic knowledge and commonsense reasoning to assist task decomposition. However, generated results must still be matched against robot skills, environmental states, and safety rules to prevent the generation of actions that exceed the robot's capabilities or violate the physical constraints of space operations. Task grounding should therefore produce structured goals whose objects, actions, preconditions, and completion criteria can be continuously checked against subsequent state estimates and robot capabilities.

Overall, research on task grounding for AI-SRO remains relatively limited, and current mainstream approaches still largely rely on predefined target recognition and fixed instruction parsing. Future research should further integrate vision-language models, space-domain knowledge, robot skill descriptions, and space-specific physical constraints to transform open-ended semantic understanding into structured task representations with explicit objects, actions, parameters, and safety boundaries. Such developments will be essential for improving the consistency of task understanding and enabling long-term autonomous operation of space robots in unknown environments.

\subsection{State Perception}

State perception aims to acquire information about the operational object, robot embodiment, and surrounding environment through multimodal sensing, including geometric, motion, and contact states. In space environments, extreme illumination, weak textures, radiation interference, and image compression can degrade visual observations, while occlusion, tumbling, and drift of noncooperative targets increase the difficulty of continuous state estimation. During close-proximity manipulation, vision alone is also insufficient for identifying contact states and target physical properties. Accordingly, state perception mainly comprises three aspects: visual enhancement, 3D reconstruction and pose-state estimation, and visuotactile fusion-based identification.

\textbf{(1) Visual Enhancement in Extreme Environments}

Alternation between direct solar illumination and deep shadow can cause image overexposure or underexposure, while radiation, sensor aging, and relative motion may introduce noise and blur. Constraints on onboard storage and communication can further result in compression artifacts. Visual enhancement should therefore not only improve image quality, but also preserve task-relevant features such as target contours, keypoints, and component boundaries. Aberdeen et al. \cite{Aberdeen2025DeepLearning_R068} combined U-Net-based image restoration with ResNet50-based pose estimation to improve pose-recognition accuracy for blurred space targets, indicating that joint optimization for downstream tasks is more suitable for SRO than simply improving perceptual image quality. Pushpakar et al. \cite{Pushpakar2013Resourcesat2_R069} investigated restoration of compression artifacts in satellite remote-sensing images and improved the preservation of edges and fine details. Existing methods are mostly designed for individual degradation types. Future approaches should incorporate space-imaging mechanisms to jointly address illumination variation, noise, blur, and compression distortion, while improving perceptual reliability through task-driven restoration and uncertainty estimation.

\textbf{(2) 3D Reconstruction and Pose-State Estimation}

3D reconstruction and pose-state estimation are used to recover target geometry and continuously estimate position, attitude, velocity, and motion trends. Objects involved in future AI-SRO tasks will often lack artificial markers and may exhibit weak textures, highly reflective surfaces, and unknown configurations. Severe occlusions may also arise from the robot manipulator and the target structure itself. These challenges require the joint use of multiview observations, temporal information, and dynamic priors. Liu et al. proposed Spacecraft-NeRF \cite{Liu2025SpacecraftNeRF_R070}, which employs neural radiance fields for high-fidelity 3D reconstruction of spacecraft. Sun et al. \cite{Sun2025NeuralRadiance_R071} introduced semantic and geometric consistency constraints to improve scene-reconstruction quality from sparse satellite imagery, enabling target semantics to remain aligned with 3D spatial structures. For dynamic unknown targets, Asri et al. \cite{Asri2025CollaborativeSwarm_R072} used multiple servicing spacecraft to share local point-cloud information and collaboratively estimate target shape and motion states. Rondao et al. proposed ChiNet \cite{Rondao2023ChiNetDeep_R073}, which integrates RGB, thermal-infrared, and temporal features to improve the continuity of relative pose estimation under complex imaging conditions. Patel et al. \cite{Patel2026GeneralPurpose_R066} combined target-motion prediction with receding-horizon planning to update interception behavior when trajectories are initially unknown and observations are limited, while Qu et al. \cite{Qu2025OcclusionAware_R067} used trajectory prediction to compensate for delayed or occluded observations and maintain alignment between observed and actual states. Space-target perception is thus evolving from single-frame pose estimation based on known models toward online reconstruction of unknown targets and multimodal temporal state estimation. Future research should further integrate target dynamics, robot motion, and observation uncertainty to maintain state prediction during short-term occlusions and communication delays while satisfying onboard real-time computation requirements.

\textbf{(3) Visuotactile Fusion-Based Identification}

In contact-rich tasks such as grasping, insertion, disassembly, and capture, visual observations are easily occluded by the manipulator and target structure, while target pose, physical properties, and contact states are difficult to infer directly from images. Contact signals in microgravity are relatively weak and may be further affected by base disturbances and structural vibrations. It is therefore necessary to combine the global geometric information provided by vision with the local interaction information obtained from force and tactile sensing. Jahanshahi \cite{Jahanshahi2025ComprehensiveReview_R075} reviewed tactile sensing technologies for space robotics and highlighted the special requirements imposed by vacuum, radiation, and temperature variations. More recently, Chen et al. \cite{Chen2026FusingTouch} developed a visual--tactile fusion method for 6D pose estimation of noncooperative spacecraft, using tactile observations to complement visually ambiguous or occluded target information. Lambert et al. \cite{Lambert2019JointInference} fused visual tracking with tactile or force/torque measurements through factor-graph inference to jointly estimate object motion, contact locations, and interaction forces, demonstrating robust state estimation even under severe visual occlusion. These studies illustrate the complementary roles of vision and contact sensing in recovering geometric and physical interaction states. Future work should establish unified state models integrating vision, force/tactile sensing, proprioception, and action histories, while actively adjusting viewpoints or exploratory contacts to jointly estimate target pose, contact state, and physical properties.

Overall, state perception for AI-SRO is gradually forming a progressive pipeline from scene restoration, to geometric and motion-state interpretation, and further to physical-property identification. The central challenge is to maintain the continuity and reliability of state estimation under extreme illumination, weak textures, severe occlusions, and complex contact conditions. By integrating multimodal information, space-dynamics priors, and uncertainty assessment, future perception systems should provide continuous and trustworthy state inputs for decision-making, planning, and action execution.

\subsection{Decision-Making and Planning}

Decision-making and planning serve as the critical link among task understanding, state perception, and action execution. Their objective is to generate operational plans that are logically valid, physically executable, and compliant with safety constraints according to task objectives and the current environmental state. Considering the hierarchical and long-horizon nature of SRO, this process mainly includes long-horizon task decomposition and planning, multi-constraint task and motion planning, and online monitoring and replanning, which respectively produce task sequences, motion trajectories, and online adjustment strategies.

\textbf{(1) Long-Horizon Task Decomposition and Planning}

On-orbit maintenance, space-facility construction, and extraterrestrial base construction typically involve multiple interdependent stages, such as target inspection, approach, grasping, transportation, assembly, and state verification. Long-horizon task planning requires decomposing a global mission objective into executable subtasks or atomic actions, while explicitly defining their preconditions, execution order, completion criteria, and resource requirements. As the number of task steps increases, the state and action search spaces expand rapidly. At the same time, an incorrect action order or an erroneous state assessment at any stage may compromise the consistency of subsequent operations and may even lead to collisions or equipment damage.

Existing studies mainly reduce the complexity of long-horizon task planning through formal methods, hierarchical architectures, and agent-based planning. Chu et al. \cite{Chu2025CollaborativePath_R077} combined linear temporal logic with control barrier functions to transform continuous lunar-base construction into multi-robot collaborative task sequences satisfying temporal, obstacle-avoidance, and energy constraints. SpaceMind \cite{Wu2026SpaceMindModular_R078} employs a modular embodied vision-language agent that integrates skill routing, tool use, and reasoning modes to autonomously organize multistage on-orbit servicing tasks and accumulate operational experience. Moser et al. \cite{Moser2022AutonomousTask_R079} incorporated stochastic task states, failure probabilities, and robot capabilities into task allocation for on-orbit assembly, allowing plans to account for execution failures, fallback actions, and task reassignment. However, these approaches may still suffer from a disconnect between high-level task plans and the feasibility of low-level skills. Future methods should therefore incorporate skill preconditions, reachability prediction, and execution-cost estimation so that the robot's physical capability to execute a planned action can be verified during task generation.

\textbf{(2) Multi-Constraint Task and Motion Planning}

A high-level task sequence can be executed only if feasible trajectories exist under the robot's kinematic, dynamic, and environmental constraints. AI-SRO therefore requires the integration of discrete task planning with continuous motion planning, while simultaneously considering obstacles, joint limits, end-effector reachability, disturbances induced by a free-floating base, momentum and energy consumption, task deadlines, and conflicts among multiple robots. In essence, this process seeks an operational plan that satisfies both task logic and physical feasibility through an iterative loop of task generation, motion validation, constraint feedback, and plan revision.

Belmonte-Baeza et al. \cite{BelmonteBaeza2026PathPlanning_R080} combined trajectory optimization with reinforcement learning, using an optimizer to generate reference trajectories satisfying dynamic constraints and a learned policy to compensate for model errors and environmental disturbances. SpaceOctopus \cite{Zhao2024SpaceoctopusOctopus_R042} formulated planning for a multi-arm space robot as a decentralized partially observable Markov decision process and employed hierarchical multi-agent reinforcement learning to coordinate the motions of different manipulators and the base. Overall, learning-based AI methods can improve adaptability in uncertain environments, whereas conventional model-based planning offers clearer constraint handling and stronger interpretability. Their integration therefore represents an important direction for solving multi-constraint motion-planning problems in AI-SRO.

\textbf{(3) Online Monitoring and Hierarchical Replanning}

Decision and planning results may become invalid during execution because of target motion, perception errors, dynamic obstacles, execution failures, or changes in mission conditions. The planned state must therefore be continuously compared with the actual state, and corrective measures such as local trajectory adjustment, skill switching, or task redecomposition should be selected according to the severity and scope of the deviation.

Existing work on online adaptation is concentrated mainly at the motion level. Fourie et al. \cite{Fourie2024ManifoldStrategies_R083} used dynamical-system modulation to adjust motion directions online according to the state of nonconvex obstacles while preserving trajectory stability. Zhou et al. \cite{Zhou2025ParallelMPPI_R084} proposed a parallel model predictive path integral method that combines multiple sampling-based planners with dynamic distance-field costs to enable real-time dynamic obstacle avoidance for robotic manipulators. Direct studies on online replanning at higher levels of SRO remain limited. A useful architecture is therefore hierarchical: minor execution deviations are handled through local trajectory adjustment; skill-execution failures trigger parameter correction or skill switching; and substantial changes in targets, resources, or mission conditions trigger high-level task redecomposition. Such a mechanism avoids invoking global replanning for every anomaly while balancing online responsiveness and global task consistency.

Overall, decision-making and planning for space robots should form a complete pipeline from task-logic decomposition, through physical-feasibility validation, to dynamic correction during execution. Existing methods can separately address long-horizon tasks, multiple constraints, and local environmental changes, but unified state representations and feedback mechanisms across different planning levels remain lacking. Future research should integrate formal models, space-dynamics priors, and data-driven methods to improve planning efficiency and online adaptability in unknown environments while preserving task logic and safety boundaries.

\subsection{Action Execution}

Action execution converts task sequences, motion paths, and target trajectories into robot commands for safe motion and physical interaction. Existing work can be organized into safe motion policy learning, manipulation policy learning under limited data and computing resources, foundation-model-driven policies, and AI-enhanced control.

\textbf{(1) Safe Motion Policy Learning in Uncertain Environments}

Space manipulators operate under uncertain geometry, occlusion, illumination variation, collision constraints, force limits, and base-disturbance requirements. IL and RL can improve online adaptability. Shao et al. \cite{Shao2026PracticalFinite_R085} developed a GAN/LSTM-based IL framework for spacecraft-mounted soft manipulators to generate motions in environments with dynamic obstacles. Su et al. \cite{Su2026ImitationLearning_R086} jointly learned a dynamical system and a Lyapunov function, providing convergence guarantees for imitation-learned motion policies. Ma et al. \cite{Ma2023ActualShape_R087} combined quadratic programming with a recurrent neural network to achieve actual-shape obstacle avoidance and trajectory tracking for redundant manipulators. These studies illustrate how learning can be embedded into structured motion-generation mechanisms rather than used as an unconstrained policy generator. Wang et al. \cite{Wang2022CollisionFree_R088} decomposed free-floating manipulator motion into high-level RRT-based planning and low-level RL tracking to obtain continuous collision-free trajectories. Al Ali et al. \cite{AlAli2024PathPlanning_R089} employed deep RL for motion planning of 6-DoF free-floating space manipulators. Tian et al. \cite{Tian2025BehaviorCloning_R090} combined behavior cloning with TD3 and used expert demonstrations to alleviate sparse-reward training in cable-driven continuum space robots. Huang et al. \cite{Huang2023ObstacleAvoidance_R091} introduced RL into the null space of a redundant space manipulator while accounting for base disturbances, enabling dynamic-obstacle avoidance without sacrificing the motion task. Li et al. \cite{Li2022ReinforcementLearning_R092} explicitly incorporated geometric and energy constraints into RL to restrict policy exploration. Collectively, these studies improve adaptability and training efficiency, but engineering deployment still requires tighter integration of kinematic/dynamic models, learned policies, and verifiable safety mechanisms such as constraint filters or control barrier functions.

\textbf{(2) Manipulation Policy Learning under Limited Data and Computing Resources}

Space manipulation tasks such as capture, insertion, disassembly, and assembly provide few real demonstrations, while trial-and-error interaction is costly. Existing studies therefore use demonstrations, prior policies, hierarchical skill representations, and multimodal observations to improve policy learning. Cao et al. \cite{Cao2023ReinforcementLearning_R040} introduced prior-policy guidance into RL for a dual-arm free-floating space robot, accelerating policy learning while preserving feasible motion behavior. Gao et al. \cite{Tian2025DemonstrationEnhanced_R096} combined learning from demonstration with RL for collaborative space multi-arm manipulation, using probabilistic movement primitives learned from demonstrations to initialize policy search before further optimization. Harris et al. \cite{Harris2022GenerationSpacecraft_R097} incorporated domain knowledge into masked RL to restrict the action space during spacecraft-operation procedure generation, illustrating how prior constraints can reduce inefficient exploration. Zhang et al. \cite{Zhang2025HRIL_R098} proposed H-RIL for modular antenna assembly, decomposing demonstrations into reusable motion skills and adapting their local composition to new assembly states. For contact-rich fine manipulation, Lang et al. \cite{Lang2025VDTFACT} proposed VDTF-ACT, which integrates visual, depth, and tactile observations within an enhanced Action Chunking with Transformer architecture for satellite-robot peg-in-socket insertion. These studies show that demonstrations, structured priors, and multimodal interaction information can substantially reduce the learning burden and improve manipulation robustness. However, most learned skills remain tied to specific tasks, robot embodiments, and interface geometries; broader transfer will require reusable skill representations and pretrained policy priors that can be efficiently adapted to new space-manipulation tasks.

\textbf{(3) Foundation-Model-Driven Action Policy Learning}

Foundation models pretrained on large-scale vision, language, and robotic data provide general semantic and behavioral priors beyond task-specific IL and RL. For action generation, VLA and WAM further couple language and visual observations with robot states and actions, enabling pretrained knowledge to be adapted to new tasks with limited demonstrations \cite{Ma2026SurveyVision_R030,Wang2026WorldAction_R031}. Current space-oriented studies, however, span different levels from multimodal cognition to direct policy generation. Foutter et al. \cite{Foutter2025SpaceLLaVA_R103} adapted a pretrained vision--language model to extraterrestrial imagery, strengthening visual-semantic understanding but without directly producing manipulation actions. Wu et al. \cite{Wu2026SpaceMindModular_R078} developed SpaceMind, which uses VLM reasoning together with skill routing and tool invocation to select and organize executable robot skills for multistage on-orbit servicing; low-level actions are still produced by external skill modules. More directly, Huang et al. \cite{Huang2026IlluminationAware_R104} fine-tuned a pretrained VLA for autonomous lunar manipulation using lightweight LoRA adaptation and illumination-aware augmentation, demonstrating foundation-policy transfer to space-specific visual conditions. These studies suggest a progression from foundation-model-assisted task reasoning toward foundation-policy action generation. Nevertheless, large-scale policies pretrained on diverse space-manipulation data have not yet emerged, and their development is constrained by scarce demonstrations, heterogeneous embodiments and action spaces, dynamics and visual domain gaps, and limited onboard computing and power resources.

\textbf{(4) AI-Enhanced Control for Difficult-to-Model Disturbances}

Action policies ultimately depend on low-level controllers capable of handling base--manipulator coupling, flexible vibration, parameter uncertainty, and variable contact. In current space-robot studies, AI is mainly used to enhance specific components of structured controllers rather than replace model-based control end to end. Huang et al. \cite{Huang2022TrackingVibration_R081} combined trajectory tracking with vibration suppression for a free-floating manipulator with flexible links and joints, while Zhang et al. \cite{Zhang2017LearningControl_R082} used model-based policy search to learn compensation policies for space robots with flexible appendages from observed trajectories. Shang et al. \cite{Shang2022DynamicModeling_R105} combined fuzzy adaptive compensation with an improved sliding-mode controller for a flexible space manipulator, using the fuzzy system to compensate uncertain nonlinear terms and disturbances caused by joint and load flexibility. Shi et al. \cite{Shi2022RobustControl_R106} proposed an optimized adaptive variable-structure controller for coordinated base--arm motion, improving robustness to inaccurate coupled dynamics through adaptive control and controller-parameter optimization. Wang et al. \cite{Wang2023RobustAdaptive_R107} integrated neural-network learning and a disturbance observer with nonsingular terminal sliding-mode control for target capture, allowing unknown post-capture coupled dynamics to be estimated and compensated online. For contact-rich operation, Liu and Chen \cite{Liu2023SpaceRobot_R108} embedded an adaptive RBF neural network into impedance control for insertion and extraction, using learned compensation to improve force regulation under model uncertainty. These works indicate that the main roles of AI in low-level SRO control are uncertainty approximation, adaptive compensation, and parameter tuning. Engineering deployment, however, still requires bounded learning, stability guarantees, and predictable real-time computation.

Overall, action execution is evolving from task-specific learned policies and model-based controllers toward reusable skills, pretrained policies, and hybrid learning--control architectures. The key challenge is to improve data efficiency and cross-task generalization without sacrificing physical feasibility, computational predictability, or safety.

\section{AI-SRO Capability Deployment and Evolution}
\label{sec_deployment}

Ground-trained AI-SRO models and learned capabilities can be transformed into practical operational capabilities only after deployment adaptation and safety validation in real space environments. Moreover, during long-duration space missions, environmental conditions, system states, and task requirements may continue to change, requiring robots to update their existing capabilities using data collected during operation. Therefore, the post-deployment stage of AI-SRO must not only address onboard execution under resource constraints and ground-to-space adaptation, but also establish continual-learning and transfer mechanisms so that operational capabilities can be reliably applied and continuously evolved.

\textbf{(1) Onboard Deployment under Resource Constraints}

The onboard deployment of AI models is first constrained by computing capability, storage, power consumption, thermal dissipation, and control-cycle requirements. In particular, foundation models such as VLMs and VLAs often contain large numbers of parameters, whereas spaceborne computing platforms place greater emphasis on radiation tolerance, low power consumption, and long-term reliability. This creates a clear mismatch between model complexity and available hardware resources. It is therefore necessary to reduce inference overhead through lightweight network design, pruning, quantization, knowledge distillation, and computational-pipeline optimization, while adapting models to processor characteristics, sensors, and control interfaces.

Existing studies have begun to address this issue. Goh et al. \cite{Goh2023SelfSupervised_R093} used self-supervised distillation to transfer knowledge from larger models to lightweight planetary-robot vision networks, while Xu et al. \cite{Xu2025LightweightNetwork_R094} developed a lightweight network for vision-based grasp-pose estimation in deep-space scenarios. Huang et al. \cite{Huang2026SLiGNet_R109} proposed SLiG-Net, which estimates grasp poses in low-illumination space environments while balancing recognition accuracy and computational efficiency. Bodmann et al. \cite{Bodmann2024EvaluatingReliability_R110} investigated the reliability of Vision Transformers in space-radiation environments and showed that even when models can run on resource-constrained computing platforms, radiation-induced computational errors may still affect output stability. Overall, current studies mainly focus on the miniaturization of individual vision or control networks, while mature solutions for onboard deployment of large-scale models such as VLAs remain lacking. In addition to model compression and inference acceleration, software and hardware redundancy, fault detection, fault-tolerant computing, and degraded operating modes should also be considered so that models can remain stable under long-term radiation exposure and progressive hardware degradation.

\textbf{(2) Ground-to-Space Adaptation and Calibration}

When models trained in terrestrial or simulated environments are deployed in space, changes in illumination, sensor noise, microgravity dynamics, structural parameters, and contact conditions may shift the state-action distribution and degrade policy performance. Ground-to-space adaptation should therefore not only expand the distributions of environmental and physical parameters during training, but also progressively calibrate models using ground experiments and real mission data to reduce discrepancies among simulation, ground testing, and actual space environments.

Existing studies mainly improve adaptation to space environments through domain adaptation, domain randomization, procedural environment generation, and staged validation. Hashimoto et al. \cite{Hashimoto2022DomainAdaptation_R099} investigated uncertainty-aware domain adaptation for 6-DoF pose estimation, while Perez-Villar et al. \cite{PerezVillar2024SpacecraftPose_R100} introduced structural losses and unsupervised intermodel-consensus adaptation for spacecraft pose estimation. Orsula et al. \cite{Orsula2022LearningGrasp_R111} trained grasping policies using procedurally generated lunar scenes and domain randomization and achieved zero-shot Sim-to-Real transfer on a real robot operating in a lunar-analogue environment. Stewart et al. \cite{Stewart2025CrossingSim2Real_R112} used curriculum learning to train control policies for the Astrobee free-flying robot and, after ground testing, deployed them on the International Space Station, providing an on-orbit demonstration of transferring RL control policies from simulation to a real microgravity environment. Together, these studies show that visual-domain alignment, broad-coverage simulation distributions, and progressive validation are complementary means of reducing ground-to-space discrepancies.

In addition, temperature variations, structural deformation, joint bias, and long-term wear may cause robot parameters to drift over time. Daş et al. \cite{Das2023ActiveLearning_R113} used Gaussian processes and active learning to select a small number of highly informative calibration points, thereby reducing the number of experiments required for autonomous recalibration of a sampling manipulator. Negi et al. \cite{Negi2025KinematicModel_R114} addressed thermal deformation and encoder bias in on-orbit manipulation by estimating kinematic parameters using contact manifolds, joint encoders, and binary contact information.

Taken together, ground-to-space transfer should combine simulation randomization, staged validation, and limited on-orbit calibration, with bounded updates and safety controllers preventing adaptation from introducing new risks.

\textbf{(3) Capability Evolution and Transfer}

During long-duration space missions, operational objects, robot states, and mission requirements may continuously change, making it difficult for policies trained only before launch to cover all future operating conditions. AI-SRO therefore needs to improve existing capabilities using experience accumulated during task execution and transfer previously acquired knowledge to new tasks, environments, and robot configurations, thereby reducing the cost of recollecting data and independently retraining policies.

At present, research directly addressing capability evolution for space robots remains limited, and existing work is mainly reflected in experience-driven skill evolution. SpaceMind \cite{Wu2026SpaceMindModular_R078} employs a Skill Self-Evolution mechanism that reflects on the outcome of each task execution and stores useful experience as reusable skill files without modifying the underlying foundation model. This approach suggests that, for space robots constrained by limited computing resources and training data, capability evolution does not necessarily have to rely entirely on model fine-tuning or incremental training; it can also be achieved through the incremental accumulation of memory, skill libraries, and experience-based knowledge.

Capability transfer is also progressing toward shared multitask policies and cross-configuration policies. Castan et al. proposed HYPER-GNC \cite{Castan2026LearningAdaptive_R115}, which uses a hypernetwork to generate shared control policies according to task descriptions, enabling a single model to cover tasks such as velocity tracking, docking, inspection, and obstacle-avoidance navigation, while adapting to new task parameters without retraining. Mishra et al. \cite{Mishra2025MultiModal_R116} employed decentralized reinforcement learning for modular lunar robots, allowing wheeled and manipulator modules to learn reusable policies independently and achieve zero-shot generalization to previously unseen robot configurations.

Continual learning and transfer should expand capability from limited on-orbit experience while controlling catastrophic forgetting, erroneous-experience accumulation, cross-configuration mismatch, and unsafe online updates.

Overall, deployment and evolution must transform ground-trained algorithms into systems that remain deployable, adaptable, and verifiable under long-duration resource and safety constraints.


\section{Future Research Directions}
\label{sec_future}

To progress from task-specific intelligence toward long-duration autonomous operation, AI-SRO research should prioritize five system-level directions.

\textbf{(1) Trustworthy Simulation and Data for Real Missions.}
High visual realism alone is insufficient for policy transfer when imaging, microgravity dynamics, flexible-body vibration, contact, and sensor degradation are coupled. Future simulators should target task-level physical fidelity and validated failure modes, while multimodal datasets should cover nominal, anomalous, and failure conditions with unified formats and evaluation protocols. Simulation and data should support not only training but also capability-boundary analysis and mission assurance.

\textbf{(2) Reliable Multimodal Cognition and State Modeling.}
Open environments around unknown spacecraft, debris, and planetary terrain require more than closed-set recognition. Future systems should integrate vision, depth, force/tactile sensing, proprioception, and space-dynamics priors to jointly represent object semantics, manipulability, geometric and motion states, contact conditions, and their uncertainty. When uncertainty becomes excessive, robots should actively change viewpoints, acquire contact information, or request external assistance rather than continue with overconfident state estimates.

\textbf{(3) Long-Horizon Safe Decision-Making under Multiple Constraints.}
Maintenance and construction require simultaneous reasoning about task order, reachability, dynamics, energy, communication, and safety. VLM-based agents such as SpaceMind demonstrate the potential of foundation models for organizing multistage tasks \cite{Wu2026SpaceMindModular_R078}, but semantically plausible plans may remain physically infeasible. Future systems should combine open-ended task reasoning with physics-based feasibility checks, formal constraints, and hierarchical recovery mechanisms that select local correction, skill switching, or global replanning according to anomaly severity.

\textbf{(4) Physically Constrained Action Generation with Limited Data.}
General robotic priors can reduce dependence on scarce space demonstrations, but embodiment, dynamics, and visual domain gaps remain substantial. High-value demonstrations, simulation, terrestrial pretraining, imitation learning, and reinforcement learning should be combined, while collision, base stability, force, and contact constraints are enforced during both training and inference. Policy confidence, safety filters, and interfaces to conventional controllers are essential so that learned actions can be verified or overridden.

\textbf{(5) Space Computing and Continual Capability Evolution.}
Foundation-model-based perception and reasoning will increasingly exceed the resources of individual robots. Future architectures should combine robot-side computation, space-edge nodes, and ground high-performance computing, supported by radiation-tolerant hardware, unified robot interfaces, and closed-loop data pipelines. Modular and reconfigurable robots can broaden physical capability \cite{Post2021ModularityFuture_R117}, while multi-robot systems can share models, skills, and experience \cite{Leitner2009MultiRobot_R118}. Demonstrations of on-orbit computing and AI platforms \cite{Gardill2023TowardsSpace_R119,Melega2025DevelopmentImplementation_R120,Xu2023AISPACE_R121} suggest a longer-term transition from intelligence embedded in individual robots toward shared space embodied-intelligence infrastructure. Continual learning within such systems must remain traceable, resource-aware, and safety-verified.

These directions emphasize that future progress will depend less on isolated algorithmic gains than on closing the loop among trustworthy data, open-world cognition, safe decision-making, physically grounded action, computing infrastructure, and long-term capability evolution.

\section{Conclusion}
\label{sec_conclusion}

This article reviewed AI-enabled space robot operations from a capability-building perspective. We summarized representative SRO scenarios and space-specific constraints, established a three-layer framework of capability foundations, capability formation, and capability deployment and evolution, and synthesized advances in simulation and data, task understanding, perception, planning, action generation, and real-system adaptation. The central challenge is not simply to improve individual AI modules, but to ensure that learned capabilities remain physically consistent, resource-feasible, transferable, and verifiably safe throughout training and deployment. Progress in trustworthy simulation, multimodal cognition and state modeling, constrained decision and policy learning, and space computing will be critical for moving space robots from task-specific automation toward long-duration, transferable, and continually evolving autonomous operation.


\bibliographystyle{IEEEtran}
\bibliography{references}

\vfill

\end{document}